\documentclass{acm_proc_article-sp}

\begin{document}

\title{Watsonsim: Overview of a Question Answering Engine}
%
%
%
%
%

\numberofauthors{3} 
%
\author{
%
%
\large{Sean Gallagher, Wlodek Zadrozny, Walid Shalaby, and Adarsh Avadhani}\\
\large{The University of North Carolina at Charlotte}\\
\large{\{sgalla19,wzadrozn,wshalaby,amanhuna\}@uncc.edu}
}

\maketitle
\begin{abstract}
The objective of the project is to design and run a system to answer Jeopardy questions, similar to Watson. In the course of a semester, we developed an open source question answering system using the Indri, Lucene, Bing and Google search engines, Apache UIMA, OpenNLP, and Weka among many additional modules. By the end of the semester, we achieved 18\% accuracy on Jeopardy questions, and work has not stopped since then.
\end{abstract}

\section{Introduction}
Given the time constraint, we prioritized our effort, starting with a minimal working subset of the functionality necessary to answer a question. Consequently, the first demonstration consisted of indexing and querying Wikipedia with Lucene, using titles as answers. From there, we added supporting passage retrieval, question classification, additional sources, evidence scoring, scoring models, and so on.

As a team, we were coordinated into smaller groups, where each group specialized in one facet of the pipeline. In total, there were teams for source acquisition, search and query generation, machine learning and scoring, and group integration.

Development started without Apache UIMA, because at the time we were not familiar with it. Midway through the semester, developers were split as to whether the project should be refactored to use UIMA but we opted to retain our existing solution. Whether this was optimal remains a significant question.


\section{Data sources}
\subsection{Offline sources}
Data collection is a major part of the question answering process, and occupied a large fraction of effort expended by the team. The greatest focus was on retrieving, cleaning, processing and indexing Wikipedia materials. These included the full texts of all Wikipedia articles from the publicly available XML dump, where redirects were removed and indexed only as synonyms to their target articles' titles. Page view statistics for 100 days of Wikipedia traffic were chosen at random by a PRNG and made into a histogram to determine page popularity, for use as a score. Two corpora of Wikipedia texts were made, where one is a subset intended to allow running the machine on underpowered machines or for those with slow internet connections who could not download the full indices.

More collections were also made available, including the full text of Wikiquotes, and the works of Shakespeare, which were indexed to match titles with works and to match  character's line with their respective names.

\subsection{Online sources}

Unlike IBM Watson, our team is not limited to offline data sources. Most notably, we use web search engines such as Google and Bing. Google query limits made mass searches of the magnitude necessary for passage retrieval impractical, even when the results were cached and every team member used their daily quotas continuously. At some point a sufficient number of results would have been reached, were it not for the issue that the queries changed continuously in an effort to improve result quality. As a result, the Google search plugin is available but not used in performance measures.

The situation with Bing is similar but less extreme. The Bing automated search quota at the time of this writing is 5000 per month, as opposed to the Google search quota of 100 per day. The longer interval available with Bing was only sufficient to allow the team to make two performance measurements over the course of the semester, but doing so required less team coordination. As a result, there are performance measurements available for Bing.

\section{Data Usage}
We used the Wikipedia reduced and full texts, along with Wikiquotes full text, and the Shakespeare full texts as soon as they were available.

Wikipedia redirects were easily integrated, but added very many generally unhelpful candidate answers. While it added 6\% binary recall (from 22\% to 28\%), it caused a 30\% drop in mean reciprocal rank (from 0.6469 to 0.3483), which dropped the overall accuracy at that point by 6\% (from 12\% to 6\%).


\section{Text Analysis}

	\subsection{Query Generation}
	There are two major question analysis pipelines, and one additional obsolete pipeline. The selection of which pipeline a question would follow was made according to its category, of which there were a small predefined set. The category was decided at the time the question is constructed, by running a set of manually developed regular expressions. Afterward, the default pipeline would choose a more specific path of analysis.
	
	The first and most common pipeline analyzes questions as factoids, or general trivia. Queries along this pipeline are weighted to favor question text found in the document content, but disfavor question text found in the title. By default, all querying engines except Google are used in the initial stage of the factoid pipeline, and Bing and Lucene querying engines are available for the later passage retrieval query stage. No configuration of the Indri search engine was found where querying speed was too slow to allow passage search; the average query in every case cost approximately 4 seconds, but over 100,000 queries were necessary for even the reduced-size performance measuring set of 100 questions, which was too much delay for rapid development.
	
	The second pipeline is more specific, targeting only fill-in-the-blank style questions. Answer-level queries are generated in generally the same manner as the factoid pipeline, with the exception that strings of underscores are removed before querying. Resulting titles are parsed to eliminate common strings, such that the only remaining text from the candidate answer is the content of the string beginning at the first blank in the question and ending at the last blank in the question. It was assumed that including the known text between blanks was acceptable. This may or may not have been the case when judged for correctness automatically, since the overlapping content between blanks may have contained strings not on the list of stopwords. Following answer extraction, passage results and candidate answers were then scored according to the same set of scorers as the factoid pipeline. This was a matter of some debate in the team since the relevancy of fill-in-the-blank passage results was likely not a function of the content of the passages. In the end, the converged pipeline was chosen, for simplicity more than any other factor.
	
	The third and least significant pipeline was designed to target quotes. The answer query was unweighted according to document titles and contents, and made no changes to the default scorers. Its difference was that the query would be directed only toward the Wikiquotes dataset, in order to reduce false positives. The solution may have been successful were it not for lack of encapsulation. It was eventually reverted, in part to allow more efficient development of the fill-in-the-blank pipeline.

	\subsection{Supporting Passage Scoring}
	Supporting passages receive a number of scores, which are intended to be indications of how much evidence the passage gives toward the correctness of the answer for which it was retrieved. Some scores are taken from the original search engines; this is the case for both offline search engines, Lucene and Indri. Lucene bases its score on Term Frequency-Inverse Document Frequency (TF-IDF) and the Vector Space Model, whereas Indri scores are based on Bayesian inference networks. \cite{Muir:LuceneSimilarity} \cite{Fisher:IndriScoring} The team expected that these models would generate substantially different results, which is supported by resulting accuracy. Using Indri raised binary recall from 13\% with only Lucene to 21\% with both engines, as measured from the first 10 results from each engine.
	
	Online search engines do not provide scores for their search results in publicly available APIs, and neither is it known how their scores are calculated. However, result ranks are still available and are included for scoring. The usefulness of these scores is diminished on the account that nearly every search will have the same minimum, maximum, and mean scores. Only queries sufficiently rare as to return no responses are meaningful in this context.
	
	Passages are also subjected to a number of n-gram comparisons. First, the passage text, candidate answer text, and question text are all filtered according to the default Lucene English analyzer, which includes stemming and stopword removal. Unigrams, bigrams, skip-bigrams, and trigrams are all taken from the each text, and the number of common n-grams is for each variety. Unigrams receive raw counts as well as frequency relative to the sum of the lengths of the texts. This suite of related scorers provided the strongest boost to ranking found in the project.
	
	 Additional scorers take an approach similar to the n-gram models, but in place of n-grams, they search for the number of common phrase subtrees taken from the Open- and CoreNLP parsers. Notably, this scorer applies to any length phrase, and the weight of the phrases found in common increases in relation to its length , because large common phrases will consist of smaller common subphrases, each of which are counted.

	\subsection{Pluggable Ranking}
	Watsonsim was originally designed for use with Mahout, and it was thought that specialized machine learning would be necessary. But early experiments involved Weka, for which some integrating scripts were made. Later it became a type of its own, a Learner, and finally was rewritten as a Researcher. Under all of these names, the function was the same, simply to convert a large fixed-length vector of scores into one floating-point score representing overall confidence in the candidate answer.
	Until the Researcher revision, the machine learning plugin was fed the entire vector of scores, and the only preprocessing to be done was to order the scores in a consistent manner. It was postulated that there were too many variables that did not represent actually distinct concepts; the scores of the first passage of evidence for a candidate answer are conceptually related to the scores for the second passage, which indicated that a single score needed to be made from both. In this revision was added a feature to take the minimum, maximum, and mean of evidence scores.
	Using Weka allowed a relatively simple way to replace one machine learning algorithm with another, where logistic regression was indicated to be the best choice. Na\"ive bayesian classifiers performed very poorly, but later results showed that Weka's default implementation of the multilayer perceptron was marginally more accurate.

\section{Dependencies}

\begin{enumerate}
\item Java JDK version 8.
\item Gradle 2 or newer for dependency management, and an Internet connection when first installing.
\item The sources database, which at the time of this writing is available online 
\item Lucene and Indri indexes, which are included as part of the extended sources.
\item A Bing account, which you must make yourself.
\item At least about 1.5GB of memory.
\item At least 50GB of free disk space, preferably more.
\end{enumerate}

\section{Performance Evaluation}

The performance tests are executed using a JUnit test that automates queries to the default pipeline, so everything needed to run the default pipeline is also needed to evaluate the performance. You will also need a way to run JUnit tests. You may use \texttt{gradle test}, or you may find Eclipse more to your liking.

\section{Current performance}

The performance tests for Watsonsim are created by taking a set of Jeopardy! questions with known correct answers, running them through the default pipeline, and collecting statistics on the resulting ranked candidate answer lists. These data were tabulated and uploaded online. The following data were taken from the peak that Watsonsim accomplished, shortly before the end of the Spring 2014 semester.


\begin{table}
\centering
\caption{Results using all sources except Google}
\smallskip
\begin{tabular}{ll}
\hline
Recall of Rank 1 & 187 (18.70\%) \\
Recall in Top 3 & 278 (27.80\%) \\
Recall in Full Candidate Answer Set & 497 (49.70\%)\\
MRR & 0.5072 \\ 
Total Questions & 1000 \\
Total Candidate Answers & 46881\\
Total Runtime & 23562.1203 \\
\hline
\end{tabular}
\end{table}

Candidate answers were marked as correct or incorrect based on whether the candidate and known correct answers both contained the same unordered set of words after filtering as done by the Lucene EnglishAnalyzer. After all candidate answers were marked as correct or incorrect, the `Recall of Rank 1' was calculated as:

\[ \frac{\mbox{count of correct answers in rank 1}}{\mbox{question count}} \]

`Recall of Top 3` is similar, with correct answers in ranks 1, 2 or 3, indexing from 1. The number of answers varied significantly; some questions had no candidate answers, and there was a limit of 90 total, 50 from Bing, 20 from Lucene, and 20 from Indri. In practice, multiple engines would return similar results, which were detected and merged. So the average was much lower, around 47 distinct candidates per question.

The Mean Reciprocal Rank was only slightly more complicated. For every question for which there was at least one correct answer, the reciprocal rank of the first correct answer was recorded. Then the mean was taken of this list of reciprocals. Questions without correct answers were eliminated from the calculation because the calculation was unclear, and because the MRR was meant as a measure of interest to the scoring team, whereas recall was a measure of interest to the search team. Mixing the two would have made collaboration more complicated. The end result is that MRR is calculated to be:

\[ \frac{\displaystyle \sum_{q \in Q}{\frac{1}{R(q)}}}{|Q|} \]
where
\begin{itemize}
\item $Q$ is the set of questions for which there exists at least one correct answer among the candidates
\item $R(q)$ is the rank of the first correct answer for question $q$
\end{itemize}

\section{Next steps}

\begin{enumerate}
\item Sources other than Wikipedia should be explored. The answers to many questions are not contained in Wikipedia, even if the search was perfect. However, such large corpora of plain, generally accurate, textual data are rare.
\item Answers should be extracted from document contents in addition to the titles. Semantic models as mentioned later may be able to assist in this process by generating important related phrases from a document.
\item Work on prismatic search is underway but not complete. Simple questions may be directly handled by sentences already existing in the corpus.
\item Scorers that require every passage to undergo NLP-based parsing are included but impractically slow. Parsing should be improved or severely limited in order to provide scores in a reasonable timeframe.
\item Unstructured Information Management Architecture (UIMA) is only weakly integrated into the Watsonsim pipelines, because the pipelines were designed before the UIMA architecture was known to the team. UIMA may be superior to the pipeline system designed in Watsonsim on two grounds. UIMA is readily horizontally scalable and compatible with multiple languages, which may make interaction with foreign libraries like Indri faster, more efficient, and more intuitive. Also, some systems are already based on it, such as Stanford CoreNLP, which was considered as an alternative option to the inconveniently unperformant dependency parser found in OpenNLP.
\item Watsonsim could benefit from more sophisticated scoring. Whereas n-gram similarities are closely scrutinized, synonymy and polysemy are completely ignored. Several scripts were written to generate distributional semantic models for Watsonsim but none were complete enough to warrant inclusion. Several projects have previously explored knowledge management through distributional semantic models, such as JoBimText which may be an excellent candidate for inclusion \cite{Biemann:JoBimText}. Word2Vec, a similar project uses distributional semantics to model relations between phrases, allowing an interesting arithmetical approach to metaphor. In such a vector space, "king" - "man" + "woman" is a very close approximation to the vector for "queen" \cite{Mikolav:Word2Vec}.
\item Offline querying with Indri is significantly more accurate than with Lucene alone. Lucene alone has 13\% binary recall in the first 10 results, whereas Indri and Lucene together have 21\% binary recall \textit{on the same data sources.} However, Indri performance is too poor in Watsonsim to be practical as a passage retrieval engine. It seems this must be the result of improper configuration, but no team member could reveal any error.
\end{enumerate}

\section{Conclusions}
Watsonsim shows that it is feasible for a small team on a deadline to create a working question answering system based on existing search technologies, online sources, natural language processing tools, and readily available machine learning toolkits.

As in many projects of this type, a great portion of the time is spent on collecting and cleaning the target data. The easily parsed format of Wikipedia as well as the accessible APIs offered by Google and Microsoft are in part responsible for reducing the time to the first demonstration. As for developing the overall pipeline, extensive documentation of Watson from IBM has proven helpful and a large influence on the design of the project.

\section{Acknowledgments}
We would like to acknowledge IBM for contributing lecture material, as well was the following students' substantial contributions:
Chris Gibson,
Dhaval Patel,
Elliot Mersch,
Jagan Vujjini,
Jonathan Shuman,
Ken Overholt,
Phani Rahul,
Varsha Devadas

%
\bibliographystyle{abbrv}

\bibliography{Watsonsim-09-2014}
%
%
\balancecolumns
\end{document}